\pdfoutput=1

\documentclass[11pt]{article}

\usepackage{acl}

\usepackage{times}
\usepackage{latexsym}

\usepackage[T1]{fontenc}

\usepackage[utf8]{inputenc}

\usepackage{microtype}
\usepackage{graphicx}
\usepackage{subfigure}
\usepackage{booktabs} 
\usepackage{multirow}

\usepackage{hyperref}

\usepackage{amsmath}
\usepackage{amssymb}
\usepackage{mathtools}
\usepackage{amsthm}
\usepackage{url} 

\usepackage[capitalize,noabbrev]{cleveref}
\usepackage{wrapfig}
\usepackage[subrefformat=parens]{subcaption}
\usepackage{etoolbox} 
\usepackage{listofitems} 
\usepackage[framemethod=TikZ]{mdframed}
\usepackage[subrefformat=parens]{subcaption}
\usepackage{tikz}
\usetikzlibrary{arrows,positioning,decorations.pathreplacing, trees}
\tikzset{>=latex} 
\colorlet{myred}{red!80!black}
\colorlet{myblue}{blue!80!black}
\colorlet{mygreen}{green!60!black}
\colorlet{mydarkred}{myred!40!black}
\colorlet{mydarkblue}{myblue!40!black}
\colorlet{mydarkgreen}{mygreen!40!black}
\tikzstyle{node}=[very thick,circle,draw=myblue,minimum size=30,inner sep=0.5,outer sep=0.6]
\tikzstyle{connect}=[->,thick,mydarkblue,shorten >=1]
\tikzset{ 
  node 1/.style={node,mydarkblue,draw=myblue,fill=myblue!20},
  node 2/.style={node,mydarkred,draw=myred,fill=myred!20},
}
\def\nstyle{int(\lay<\Nnodlen?min(1,\lay):2)} 

\theoremstyle{plain}

\theoremstyle{definition}

\theoremstyle{remark}

\usepackage[textsize=tiny]{todonotes}

\newcommand{\transpose}{^{\mathrm{T}}}

\title{Multi-label Learning with Random Circular Vectors}

\author{
  Ken Nishida\thanks{~~Email:~z301067a@gmail.com} \\
  Hokkaido University \\
  \And
  Kojiro Machi \\
  Hokkaido University
  \And
  Kazuma Onishi \\
  Hokkaido University \\
  \AND
  Katsuhiko Hayashi \\
  The University of Tokyo \\
  \And
  Hidetaka Kamigaito \\
  Nara Institute of Science and Technology 
}

\begin{document}
\maketitle
\begin{abstract}
\label{sec:abstract}
The extreme multi-label classification~(XMC) task involves learning a classifier that can predict from a large label set the most relevant subset of labels for a data instance. 
While deep neural networks~(DNNs) have demonstrated remarkable success in XMC problems, the task is still challenging because it must deal with a large number of output labels, which make the DNN training computationally expensive.
This paper addresses the issue by exploring the use of random circular vectors, where each vector component is represented as a complex amplitude.
In our framework, we can develop an output layer and loss function of DNNs for XMC by representing the final output layer as a fully connected layer that directly predicts a low-dimensional circular vector encoding a set of labels for a data instance.
We conducted experiments on synthetic datasets to verify that circular vectors have better label encoding capacity and retrieval ability than normal real-valued vectors.
Then, we conducted experiments on actual XMC datasets and found that these appealing properties of circular vectors contribute to significant improvements in task performance compared with a previous model using random real-valued vectors, while reducing the size of the output layers by up to 99\%.
\end{abstract}

\section{Introduction}
\label{sec:introduction}
Extreme multi-label classification (XMC) problems arise in various domains, such as product recommendation systems~\cite{jain2016extreme}, labeling large encyclopedia~\cite{dekel2010multiclass,partalas2015lshtc}, instance-level image recognition~\cite{deng2010does,ridnik2021imagenet21k} and natural language generation~\cite{mikolov2013efficient}.
The XMC task involves learning a classifier which can predict from a large label set the most relevant subset of labels for a data instance.
Recent work has focused on deep neural network~(DNN) models~\citep{liu2017deep,you2019attentionxml,chang2020taming,zhang2021fast,dahiya2023ngame,jain2023renee} that deliver task performances superior to those of early approaches using linear predictors~\citep{babbar2017dismec,prabhu2018parabel}.

While DNN models have brought great performance improvements, the XMC task still remains a challenge mainly due to the extremely large output space.
Since a large number of output labels make it difficult to train DNN models efficiently, various methods for improving training efficiency have been proposed~\citep{khandagale2020bonsai,wydmuch2018no,jiang2021lightxml,ganesan2021learning}.
Among the previous studies, \citet{ganesan2021learning} presented a promising method that employs random real-valued vectors for reducing the output layer size of DNN models.
In this approach, a high-dimensional output space vector is replaced with a low-dimensional random vector encoding the relevant label information for a data instance.
Then, DNN models are trained to predict the label-encoded vector directly.
After the model generates a vector, it can be checked approximately whether a label is encoded in it or not through a vector comparison using the cosine similarity between the output vector and a vector that the label is assigned to.
The basic idea of the label encoding and retrieval framework relies on the theory of Holographic Reduced Representations~\citep{plate1995holographic}, which was developed in the cognitive neuroscience field.

However, random real-valued vectors do not have sufficient ability for representing data instances that belong to many class concepts.
As our experiments in \S~\ref{sec:proposed_method} show, the label retrieval accuracy decreases markedly as the number of class labels encoded in a vector increases.
To alleviate the issue, this paper presents a novel method that uses {\it circular} vectors instead of real-valued vectors.
Each element of a circular vector takes a complex amplitude as its value; i.e., the vector element is represented by an angle ranging from $-\pi$ to $\pi$.
Since an angle can be represented by a real value, the memory cost for the circular vector representation is the same as that for a normal real-valued vector.
In spite of this fact, surprisingly, circular vectors have better label encoding and retrieval capacities than real-valued vectors.
One of the challenges in applying circular vectors to DNN models is how to adapt the output layer to a circular vector.
In \S~\ref{sec:network_architecture}, we describe our neural network architecture that uses circular vectors in the output layer.
Our experimental results on XMC datasets show that our method based on circular vectors significantly outperforms a previous model using real-valued vectors, while reducing the size of the output layers by up to 99\%.

\section{Previous Study: Learning with Holographic Reduced Representations}
\label{sec:previous_study}
Several vector symbolic architectures have been developed in the field of cognitive neuroscience, including Tensor Product Representations~\citep{smolensky1990tensor}, Binary Spatter Code~\citep{Kanerva1996BinarySO}, Binary Sparse Distributed Representations~\citep{917565}, Multiply-Add-Permute~\citep{gayler2004vector}, and Holographic Reduced Representations (HRR)~\citep{plate1995holographic}.
Among them, HRR is a successful architecture for distributed representations of compositional structures.
To model complex structured prediction tasks in a vector space that involve key-value stores, sequences, trees and graphs, many prior studies have explored how to use HRR in various machine learning frameworks; Recurrent Neural Networks~\citep{plate1992holographic}, Tree Kernels~\citep{zanzotto2012distributed}, Knowledge Graph Representation Learning~\citep{nickel2015holographic,hayashi-shimbo-2017-equivalence}, Long-short Term Memory Networks~\citep{DBLP:journals/corr/DanihelkaWUKG16}, Transformer Networks~\citep{alam2023recasting}, and among others. 
In particular, \citet{ganesan2021learning} presented a general framework based on the HRR architecture for efficient multi-label learning of DNN models.
To clarify the motivation of our study, we will review the framework in more detail in the following subsections.

\subsection{Holographic Reduced Representations~(HRR)}
In the HRR architecture, terms in a domain are represented by real-valued vectors.
Here, we assume that each vector is independently sampled from a Gaussian distribution $\mathcal{N}(0,\mathbf{I}_d \cdot d^{-1})$, where $d$ is the vector dimension size and $\mathbf{I}_d$ is the $d\times d$ identity matrix.
To bind an association of two terms represented by vectors $\mathbf{a}$ and $\mathbf{b}$, respectively, HRR uses circular convolution, denoted by the mathematical symbol $\otimes$:
\begin{equation}
    \mathbf{a} \otimes \mathbf{b} = \mathcal{F}^{-1} (\mathcal{F}(\mathbf{a}) \odot \mathcal{F}(\mathbf{b}))
    \label{hrr_binding}
\end{equation}
where $\odot$ is element-wise vector multiplication.
Note that the circular convolution can be computed by using a fast Fourier transform~(FFT) $\mathcal{F}$ and inverse FFT $ \mathcal{F}^{-1}$, but they require $\mathcal{O}(d\log d)$ computation time.
Given several associations $\mathbf{a} \otimes \mathbf{b}$, $\mathbf{c} \otimes \mathbf{d}$ and $\mathbf{e} \otimes \mathbf{f}$,
the vectors can be superposed to represent their combination: $\mathbf{S}=(\mathbf{a} \otimes \mathbf{b}) \oplus (\mathbf{c} \otimes \mathbf{d}) \oplus (\mathbf{e} \otimes \mathbf{f})$, where the ``superposition'' operator $\oplus$ is just normal vector addition $+$.
The HRR architecture also provides the inversion operation $\dag$:
\begin{equation}
 \mathbf{a}^{\dag}
  = \mathcal{F}^{-1}(\frac{1}{\mathcal{F}(\mathbf{a})}).
  \label{hrr_inverse}
\end{equation}
The inversion operation can be used to perform ``unbinding''.
For an example, it allows the reconstruction of a noisy version of $\mathbf{d}$ to be recreated from the memory $\mathbf{S}$ and a cue $\mathbf{c}$: $\mathbf{S}\otimes\mathbf{c}^{\dag}\approx\mathbf{d}$.
Finally, the ``similarity'' operation is defined as the dot-product $\mathbf{a}\transpose\mathbf{b}$.
Using the similarity operation, we can check approximately whether $\mathbf{a}$ exists in a memory $\mathbf{S}$ if $\mathbf{S}\transpose\mathbf{a}\approx1$ or not present if $\mathbf{S}\transpose\mathbf{a}\approx0$.

\subsection{Multi-label Learning with HRR}
\label{subsec:learning_with_HRR}
\citet{ganesan2021learning} introduced a novel method using HRR for reducing the computational complexity of training DNNs for XMC tasks.
Let $L$ be the number of class labels in an XMC task.
The basic idea behind the approach of \citep{ganesan2021learning} is quite intuitive; for efficient DNN training, an $L$-dimensional output (teacher) vector is replaced with a $d$-dimensional real-valued vector encoding the relevant label information for a data instance.
By assuming $d\ll L$, we can dramatically reduce the output layer size of the DNN model.

In this approach, each class label $y$ is assigned to a $d$-dimensional vector $\mathbf{c}_y\in\mathbb{R}^{d}$. Then, the label information for a data instance $x$ is represented as a {\it label vector} $\mathbf{S}_x\in \mathbb{R}^{d}$:
\begin{equation}
    \mathbf{S}_x = \bigoplus_{p \in \mathcal{Y}_x} \mathbf{p} \otimes \mathbf{c}_p
    \label{label_represenration}
\end{equation}
where $\mathcal{Y}_x$ denotes the set of class labels that $x$ belongs to and $\mathbf{p}\in\mathbb{R}^{d}$ represents the positive class concept.\footnote{We can encode information on negative labels into a label vector as well as positive ones, but as shown in \citep{ganesan2021learning}, the negative label information does not contribute to improving XMC task performance. Thus, in this paper, we will omit discussion on negative labels for notational brevity.}
To train a DNN model $f(\mathbf{x})$ that generates $\hat{\mathbf{ S}}_x\in\mathbb{R}^{d}\approx\mathbf{S}_x$, 
\citet{ganesan2021learning} define a loss function:
\begin{equation}
    loss = \sum_{p\in \mathcal{Y}_x} (1- sim{(( \hat{\mathbf{S}}_x\otimes\mathbf{p}^{\dag}),\mathbf{c}_p})).
    \label{eq:hrr_loss_function}
\end{equation}
To prevent the model from maximizing the magnitudes of the output vectors, \citet{ganesan2021learning} used the cosine similarity as $sim(\cdot,\cdot)$, which is a normalized version of the dot product that ranges from -1 to 1.
In the inference phase, labels can also be ranked according to the cosine similarity computed by $sim(\hat{\mathbf{S}}_x\otimes\mathbf{p}^{\dag},\mathbf{c}_p)$ for each label $p$.
Moreover, \citet{ganesan2021learning} introduced a novel vector {\it projection} method to reduce the effect of the variance of the similarity computation:
\begin{equation}
    \pi(\mathbf{x}) = \mathcal{F}^{-1} \left(\dots,\frac{\mathcal{F}(\mathbf{x})_j}{|\mathcal{F}(\mathbf{x})_j|},\dots\right).
  \label{eq:hrr_projection}
\end{equation}
Here, each HRR vector $\mathbf{x}$ is initialized with $\mathbf{x}\stackrel{\mathrm{d}}{=}\pi\left(\mathcal{N}(0,\mathbf{I}_d \cdot d^{-1})\right)$, which ensures each element of the vector in the frequency domain is unitary; i.e., the complex magnitude is one.

\section{Multi-label Representations with Circular Vectors}
\label{sec:proposed_method}
In this section, we show through experiments that random real-valued vectors actually do not have sufficient ability for representing data
instances that belong to many classes.
The reason is mainly due to the projection operation in Equation~\ref{eq:hrr_projection}.
As described in \S~\ref{sec:previous_study}, the projection operation was proposed as a way to reduce the effect of the variance of the similarity computation, but each element of the superposition between two normalized vectors via the projection is no longer unitary. Thus, the effect of the projection decreases when a label vector encodes more class labels.
To alleviate the issue, we developed a simple alternative that forces all vector elements to be unitary in the complex domain even after the superposition operation. 
We describe the details in the following subsection.

\subsection{HRR with Circular Vectors}
\begin{figure}[t]
\centering
\begin{minipage}{\textwidth}
\hspace{1.5cm} 
\begin{tikzpicture}[scale=0.4]
    \draw[thick,->] (0,-5) -- (0,5)node[left]{Im};
    \draw[thick,->] (-5,0) -- (5,0)node[below]{Re};

    \draw (4,0) arc (0:360:4);

    \draw[red] (0,0) -- (3.3,2.4) node[circle,red,fill,inner sep=2pt]{};
    \draw[red] (3.3,2.4) -- (4.8,2.4) node[right]{$(\cos \phi,\sin \phi)$};

    \node[below right] at (4,1.2) {$1$};
    \node[above left] at (1,4.5) {$1$};
    \node[below left] at (-4,0) {$-1$};
    \node[below right] at (0,-4) {$-1$};

    \draw (1,0) arc (0:38:1)node[midway,above right= -0.2cm and 0.0cm]{$\phi$};
\end{tikzpicture}
\end{minipage}
\caption{The unit circle in the complex plane with coordinates. The angle $\phi$ represents an element of the circular vector $\bar{\phi}$.}
\label{fig:unit_circle}
\end{figure}
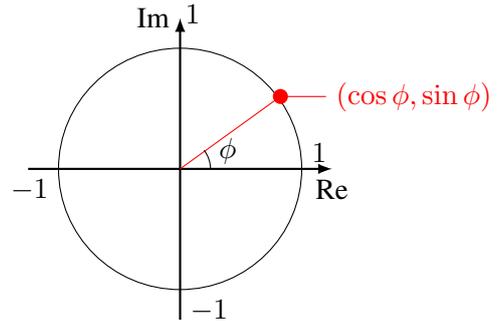

\begin{table*}[t]
  \centering
  \small
  \begin{tabular}{lll}
    \hline
    Operation & Real-valued~\citep{ganesan2021learning} &  Circular  \\ \hline
    vector      & $ {\mathbf{x}} = [x_0 ,\dots , x_{d-1}]$ & $\bar{{\phi}} = [\phi_0 , \dots , \phi_{d-1}]$ \\
    random vector  & $\mathbf{x}\stackrel{\mathrm{d}}{=} \pi\left(\mathcal{N}(0,\mathbf{I}_d \cdot d^{-1})\right)$ & $\phi_j \stackrel{\mathrm{d}}{=} \mathcal{U}(-\pi, \pi)$  \\
    binding     & $\mathbf{x} \otimes \mathbf{y} = \mathcal{F}^{-1} (\mathcal{F}(\mathbf{x}) \odot \mathcal{F}(\mathbf{y}))$ & $\bar{\phi} \otimes \bar{\theta} = [(\phi_0+\theta_0) \bmod 2\pi, \ldots ,(\phi_{d-1}+\theta_{d-1}) \bmod 2\pi]$  \\
    unbinding    & $\mathbf{x}\otimes \mathbf{y}^{\dag} = \mathbf{x}  \otimes\mathcal{F}^{-1}(\frac{1}{\mathcal{F}(\mathbf{y})}) $ & $\bar{\phi} \otimes \bar{\theta}^\dag = -\bar{\theta} \otimes \bar{\phi}$ \\
    similarity & $sim(\mathbf{x},\mathbf{y})=\mathbf{x}\transpose \mathbf{y}$ & $sim(\bar{\phi}, \bar{\theta}) = \frac{1}{d}\sum_{j}\cos(\phi_j - \theta_j)$   \\
    superposition    & $\mathbf{x} \oplus \mathbf{y} = \mathbf{x} + \mathbf{y}$ & $\bar{\phi} \oplus \bar{\theta} = [\angle(e^{i \cdot {\phi}_0} + e^{i \cdot
    {\theta}_0}), \dots ,\angle(e^{i \cdot {\phi}_{d-1}} +{e}^{i \cdot {\theta}_{d-1}})]$\\
    \hline
  \end{tabular}
  \caption{Comparison of HRR operations on real-valued and circular vectors.}
  \label{tab:operations}
\end{table*}
Our idea is to use {\it circular} vectors instead of real-valued vectors.
Circular vectors have a complex amplitude~(see Figure~\ref{fig:unit_circle}), which can be represented by a real value $\phi$ ranging from $-\pi$ to $\pi$.
However, to force all vector elements to be unitary after any operations, we require a special HRR system for circular vectors.
In this paper, we borrow the concept of a circular HRR~(CHRR) system from \citep{plate2003holographic}.

Table~\ref{tab:operations} compares the HRR operations of the standard and circular systems.
For circular vectors, each element must be sampled from a uniform distribution $\mathcal{U}(-\pi, \pi)$ over $(-\pi, \pi]$.
The binding $\otimes$ and inversion $\dag$ of CHRR are implemented with the standard vector arithmetic operations like addition and subtraction.
The similarity of two circular vectors can be simply determined from the sum of the cosines of the
differences between angles.
On the other hand, superposition is somewhat tricky because in general the sum of unitary complex values does not lie on the unit circle.
For each pair of elements $\phi_j$ and $\theta_j$ of two circular vectors $\bar{\phi}$ and $\bar{\theta}$, the result of superposition is $\angle(e^{i \cdot {\phi}_j} + e^{i \cdot
    {\theta}_j})$.
Here, $\angle(v)$ extracts an angle of a complex value $v$ and discards the magnitude of $v$.
Since all of these operations do not affect the unitary property of circular vectors, we no longer need the projection normalization process.
Our framework also has an advantage in computational cost; we can avoid the FFT and inverse FFT operations, which take $\mathcal{O}(d\log d)$ computation time.

\subsection{Retrieval Accuracy Experiment}
\label{subsec:Retrieval_experiment}

\begin{figure}[t]
    \centering
    \begin{minipage}[b]{0.48\linewidth} 
        \centering
        \includegraphics[width=\linewidth]{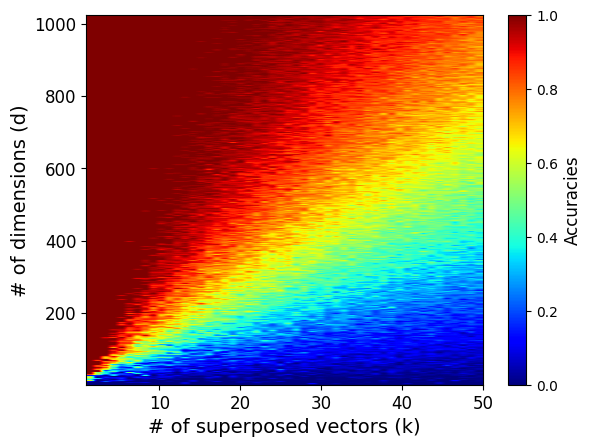}
        \vspace{-3mm}
        \caption*{(a) HRR(w/Proj)}
    \end{minipage}
    \hfill
    \begin{minipage}[b]{0.48\linewidth} 
        \centering
        \includegraphics[width=\linewidth]{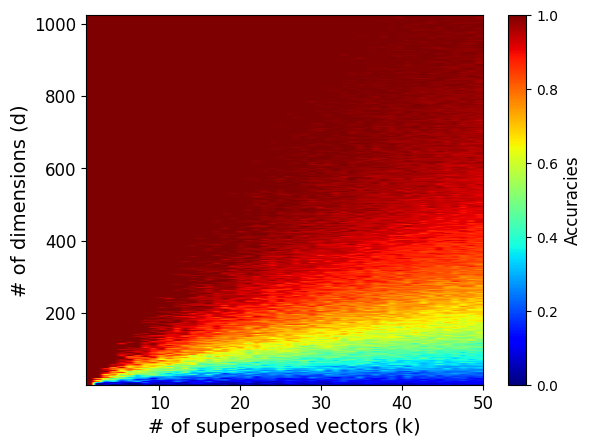}
        \vspace{-3mm}
        \caption*{(b) CHRR}
    \end{minipage}
    \caption{Retrieval accuracies of HRR(w/Proj) and CHRR.
    The number of dimensions $d$ was $1,\dots,1024$ and the number of positive classes $k$ was $1,\dots,50$.
    }
    \vspace{-3mm}
    \label{fig:Heatmap}
\end{figure}

We experimentally demonstrated CHRR's capacity by comparing its retrieval accuracy with that of HRR.
The experiment attempted to verify how accurately the positive class vector can be retrieved from a memory vector.
For a data instance $x$, let $\mathbf{c}_p$ be a vector for a positive class $p$ to which $x$ belongs, and let $\mathbf{p}$ be a vector for the positive class concept label.
The binding and superposition operations allow us to represent all positive classes for $x$ as $\mathbf{R}$:
\begin{equation} 
    \mathbf{R} = \bigoplus_{p \in \mathcal{Y}_x} (\mathbf{p} \otimes \mathbf{c}_p).
    \label{eq:3_1_R}
\end{equation}
In the experiment, we generated a database consisting of $N=1,000$ random $d$-dimensional vectors ($\mathbf{c}_j \in \mathbb{R}^d$, for all $j \in[1,\dots,N]$).
Then, to create $\mathbf{R}$, we randomly selected $k$ vectors from the database to be $\mathbf{c}_p$ and one vector to be $\mathbf{p}$.
As shown in Equation~\ref{eq:3_1_R}, the $k$ associations can be superposed to represent $\mathbf{R}$.
To retrieve $\mathbf{c}_p$ from $\mathbf{R}$, we used the unbinding operation to decode a noisy version of the vector $\mathbf{c}_p$ from $\mathbf{R}$, as $\hat{\mathbf{c}}_p=\mathbf{R} \otimes \mathbf{p}^\dagger$. 
For each $j\in[1,\dots,N]$, we computed the similarity $s_j = sim(\hat{\mathbf{c}}_p,\mathbf{c}_j)$ between the decoded vector $\hat{\mathbf{c}}_p$ and the individual vector $\mathbf{c}_j$.
After that, we compiled the top-$k$ label list according to the similarity scores $s_j$.
To evaluate the retrieval accuracy, we measured the percentage of class labels in the list, whose vectors were encoded into the memory $\mathbf{R}$.
By varying the number of dimensions $d = 1,\dots,1024$ and the number of binding pairs $k = 1,\dots,50$, we plotted the accuracies as a heat-map~(Figure~\ref{fig:Heatmap}, where
warmer colors indicate higher accuracy).\footnote{\citet{Schlegel_2021} also demonstrated that CHRR has a higher retrieval capacity compared with HRR. Yet, they used all distinct vectors: $\mathbf{R}=(\mathbf{a} \otimes \mathbf{b}) \oplus (\mathbf{c} \otimes \mathbf{d})$, and did not use a fixed $\mathbf{p}$: $\mathbf{R}=(\mathbf{p} \otimes \mathbf{a}) \oplus (\mathbf{p} \otimes \mathbf{b})$. Therefore, we changed their experimental settings to fit the XMC learning with HRR.}
The results clearly show that CHRR has better retrieval accuracies than those of HRR.
Moreover, the larger the number of superposed vectors ($k$) is, the bigger the performance difference between CHRR and HRR becomes.
Hence, this tendency indicates that CHRR is more suitable than HRR for encoding many labels.

\subsection{Variance Comparison Experiment}
\label{subsec:Variance_experiment}
\begin{figure}[t]
    \centering
    \begin{minipage}[b]{0.48\linewidth} 
        \centering
        \includegraphics[width=\linewidth]{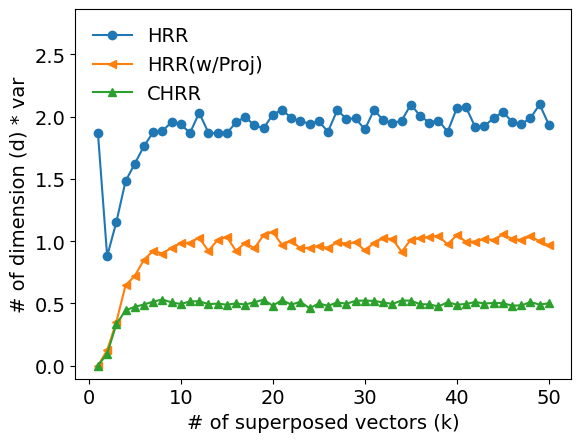}
        \vspace{-3mm}
        \caption*{(a) Variance}
    \end{minipage}
    \hfill
    \begin{minipage}[b]{0.48\linewidth} 
        \centering
        \includegraphics[width=\linewidth]{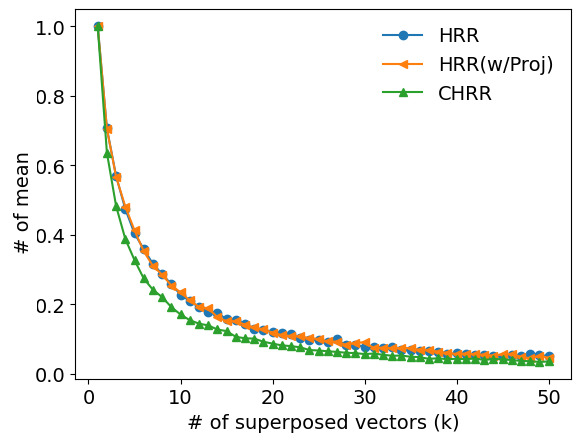}
        \vspace{-3mm}
        \caption*{(b) Mean}
    \end{minipage}
    \caption{
    Variance and mean of the similarities of HRR, HRR(w/Proj), and CHRR.
    We fixed the number of dimensions $d$ to $400$ and varied the number of positive classes $k$ from $1$ to $50$.
    }
    \vspace{-3mm}
    \label{fig:Simalarity_Mean_Variance}
\end{figure}

In ~\S~\ref{subsec:Retrieval_experiment}, we confirmed that CHRR exhibits superior retrieval ability to HRR.
There is a possibility that the CHRR's similarity operation reduces the variance more than the projection does.
The experiment reported below was conducted to check the numerical stability of the CHRR's similarity operation.
To create $\mathbf{R}$ as Equation~\ref{eq:3_1_R}, we generated $k$ random vectors $\mathbf{c}_p$ and $\mathbf{p}$.
We extracted a noisy version of $\mathbf{c}_p$ from ${\mathbf{R}}$ as $\hat{\mathbf{c}}_p=\mathbf{R} \otimes \mathbf{p}^\dagger$.
For each $j\in[1,\dots,k]$, we measure the similarity between $\mathbf{\hat{c}_p}$ and $\mathbf{c}_j$ as $s_j=sim(\mathbf{\hat{c}}_p,\mathbf{c}_j)$.
We plotted the variances and means of the similarities in Figure~\ref{fig:Simalarity_Mean_Variance}~(a) and (b), respectively.
We fixed the number of dimensions $d$ to $400$ and varied the number of binding pairs $k = 1,\dots,50$.
Our experiments compared three methods, CHRR, HRR proposed in~\citep{plate1995holographic}, and HRR with the projection of \citep{ganesan2021learning}~(HRR(w/Proj)).

Figure~\ref{fig:Simalarity_Mean_Variance}~(a) shows that as $k$ increases, the variances of all methods tend to converge.
However, while the variance converges, the mean also decreases near zero, as shown in Figure~\ref{fig:Simalarity_Mean_Variance}~(b).
Therefore, as the number of superposed vectors $k$ increases, the impact of variance becomes relatively larger.
Regarding the variance, we can see the need for the projection, since the HRR(w/Proj) is more suppressed than the original HRR.
Yet, we found that CHRR is most suppressed; that is, CHRR is more numerically stable than HRR(w/Proj).
As for the mean, the three methods had roughly comparable performances.
Although the mean approached zero as $k$ increased, this is not a problem in using similarity for compiling a ranking list of labels.

\section{Neural Network Architecture}
\label{sec:network_architecture}
One of the challenges in adapting CHRR to XMC tasks is how to adapt the output layer of DNN models to a circular vector because it has a cyclic feature; i.e., $\theta = 2 \pi n \times \theta$, where $ 
n \in \mathbb{Z}$.
To meet it, we developed a neural network for predicting angles that considers the cyclic feature during the training.
The key idea was to represent the output in Cartesian coordinates, which can uniquely represent a point on a unit circle. 
Then, we converted the output into polar coordinates to obtain angles.


\subsection{Architecture for Circular Vectors}
We used fully connected (FC) networks in all of the experiments.
They were each composed of a $F$-dimensional input layer, two $h$-dimensional hidden layers with ReLU activation~\citep{agarap2018relu}, and a $d'$-dimensional output layer.
That is, they had the same architecture except for the output layer.

We selected two baselines from \citet{ganesan2021learning} by using the FC networks.
The first baseline had $L$ output nodes and each node is used to binary classification (we refer to it below as FC).
The second baseline was the method using HRR as described in \S~\ref{subsec:learning_with_HRR}.
It had $d$ output nodes (we refer to it below as HRR).

Our network for CHRR represented a pair of the outputs as a point on a unit circle on Cartesian coordinates; i.e., $(\cos \phi, \sin \phi)$, as shown in Figure~\ref{fig:unit_circle}.
Then we converted the point into polar coordinates $(1, \phi)$, and used $\phi$ as an element of the predicted label vector.
Let $\hat{\mathbf{s}} \in \mathbb{R}^{2d}$ be the raw output vector, and $\hat{\mathbf{S}} \in \mathbb{C}^{d}$ be the converted circular vector.
We represented $d$ pairs from $\hat{\mathbf{s}}$ in Cartesian coordinates as $a_i = (x_i, y_i)$.
Then, we normalized them to satisfy $\|a_i\| = 1$.
Although there was a similar work for an angle prediction using a neural network~\citep{Heffernan2015angle_estimation}, they used $\arctan{\frac{y}{x}}$ for the conversion whose range was limited to $\left[\frac{-\pi}{2}, \frac{\pi}{2}\right]$.
Instead, we used the atan2 function~\citep{organick1966atan2}, which can convert a $(x, y)$ point to a corresponding angle $(-\pi, \pi]$.
Finally, we adapted the atan2 to $a_i$ to obtain $\hat{S}_i$.
We named this method as CHRR.

\subsection{Impact of Model Architecture}
\label{subsec:model_ablation}
Because the number of the output nodes of CHRR ($2d$) is twice as that of HRR ($d$), the total model size of CHRR also increases.
Therefore, we conducted two different experiments using the same model size as HRR (see \S~\ref{subsec:results_and_discussion} for the results).
The first experiment changed the network architecture of CHRR.
Figure~\ref{fig:architecture} compares the architectures of CHRR and the changed model (CHRR-Half) to illustrate the impact of halving the hidden and output layer sizes on model performance. This adjustment ensures that CHRR-Half has the same number of parameters as HRR, allowing for a fair comparison.
We made CHRR-Half by splitting the second hidden layer's nodes and output nodes of CHRR in half.
This resulted in two sets of $\frac{h}{2}$ hidden nodes and $d$ output nodes.
Then we connected one set of hidden nodes to one set of output nodes, and the other set of hidden nodes to the other set of output nodes.
As a result, $2 \times (\frac{h}{2} \times d) = h \times d$ parameters were obtained, which equals the number of parameters between the second hidden layer and the output layer in HRR.
The results of the experiment in \S~\ref{subsec:results_and_discussion} showed no significant difference in performance between CHRR and this model.
Therefore, the increase in the model size is not a big issue.
In the second experiment, to demonstrate the advantage of the proposed architecture against naive implementation, we used the same network architecture as HRR, and mapped the real-valued outputs to angles with activation functions.
We tried two activation functions, $\sin$ and $\tanh$ to map the outputs to $[-1, 1]$; then the output was multiplied by $\pi$ to obtain $(-\pi, \pi]$ outputs.
We named these models as CHRR-sin and CHRR-tanh.
Both showed more modest levels of performance compared with CHRR.

\begin{figure}[t]
    \centering
    \begin{minipage}{0.37\linewidth}
        \centering
        \resizebox{\linewidth}{!}{
\begin{tikzpicture}[x=2.4cm,y=1.2cm]
  \readlist\Nnod{4,4} 
  \readlist\Nstr{h,2d} 
  \readlist\Cstr{h^{(1)},y} 
  \def\yshift{0.80} 
  
  \foreachitem \N \in \Nnod{
    \def\lay{\Ncnt} 
    \pgfmathsetmacro\prev{int(\Ncnt-1)} 
    \ifnum \lay=1
        \readlist\Idxs{0,\frac{\Nstr[\lay]}{2} - 1, \frac{\Nstr[\lay]}{2}}
    \else
        \readlist\Idxs{0,d - 1, d}
    \fi
    
    \foreach \i [evaluate={\c=int(\i==\N); \cc=int(\i==1); \y=\N/2-\i--\cc*\yshift-\c*\yshift;
                 \x=\lay; \n=\nstyle;
                 \index=(\i==1?1:"\Nstr[\n]");}] in {1,...,\N}{ 
      \ifnumcomp{\i}{<}{\N}{
        \node[node \n] (N\lay-\i) at (\x,\y) {$\strut\Cstr[\n]_{\Idxs[\i]}$};
      }
      
      \ifnum \i=\N
        \node[node \n] (N\lay-\i) at (\x,\y) {$\strut\Cstr[\n]_{\Nstr[\n]-1}$};
      \fi

      \ifnumcomp{\lay}{>}{1}{ 
        \foreach \j in {1,...,\Nnod[\prev]}{ 
          \draw[white,line width=1.2,shorten >=1] (N\prev-\j) -- (N\lay-\i);
          \draw[connect] (N\prev-\j) -- (N\lay-\i);
        }
        \ifnum \lay=\Nnodlen
        \fi
      }
      
    }

    \path (N\lay-2) --++ (0,1+\yshift) node[midway,scale=1.6] {$\vdots$};
    \path (N\lay-\N) --++ (0,1+\yshift) node[midway,scale=1.6] {$\vdots$}; 
  }

 \node[above,align=center,mydarkblue] at (1,2.3) {Hidden\\[-0.2em]layer};
 \node[above,align=center,mydarkred] at (2,2.3) {Output\\[-0.2em]layer};.

\end{tikzpicture}
        }
        \caption*{(a) CHRR}
    \end{minipage}
    \hspace{1cm} 
    \begin{minipage}{0.37\linewidth}
        \centering
        \resizebox{\linewidth}{!}{
\begin{tikzpicture}[x=2.4cm,y=1.2cm]
  \readlist\Nnod{4,4} 
  \readlist\Nstr{h,2d} 
  \readlist\Cstr{h^{(1)},y} 
  \def\yshift{0.80} 
  
  \foreachitem \N \in \Nnod{
    \def\lay{\Ncnt} 
    \pgfmathsetmacro\prev{int(\Ncnt-1)} 
    \ifnum \lay=1
        \readlist\Idxs{0,\frac{\Nstr[\lay]}{2} - 1, \frac{\Nstr[\lay]}{2}}
    \else
        \readlist\Idxs{0,d - 1, d}
    \fi
    
    \foreach \i [evaluate={\c=int(\i==\N); \cc=int(\i==1); \y=\N/2-\i--\cc*\yshift-\c*\yshift;
                 \x=\lay; \n=\nstyle;
                 \index=(\i==1?1:"\Nstr[\n]");}] in {1,...,\N}{ 
          \ifnumcomp{\i}{<}{\N}{
            \node[node \n] (N\lay-\i) at (\x,\y) {$\strut\Cstr[\n]_{\Idxs[\i]}$};
          }
          
          \ifnum \i=\N
            \node[node \n] (N\lay-\i) at (\x,\y) {$\strut\Cstr[\n]_{\Nstr[\n]-1}$};
          \fi

      \ifnumcomp{\lay}{>}{1}{ 
        \foreach \j in {1,...,\Nnod[\prev]}{ 
            \ifnumcomp{\i}{<}{3}{
                \ifnumcomp{\j}{<}{3}{
                    \draw[white,line width=1.2,shorten >=1] (N\prev-\j) -- (N\lay-\i);
                    \draw[connect] (N\prev-\j) -- (N\lay-\i);
                }{}
            }{}
            \ifnumcomp{\i}{>}{2}{
                \ifnumcomp{\j}{>}{2}{
                    \draw[white,line width=1.2,shorten >=1] (N\prev-\j) -- (N\lay-\i);
                    \draw[connect] (N\prev-\j) -- (N\lay-\i);
                }
            }{}
        }
        \ifnum \lay=\Nnodlen
        \fi
      }{}
      
    }

    \path (N\lay-2) --++ (0,1+\yshift) node[midway,scale=1.6] {$\vdots$};
    \path (N\lay-\N) --++ (0,1+\yshift) node[midway,scale=1.6] {$\vdots$}; 
  }
  
 \node[above,align=center,mydarkblue] at (1,2.3) {Hidden\\[-0.2em]layer};
 \node[above,align=center,mydarkred] at (2,2.3) {Output\\[-0.2em]layer};
  
\end{tikzpicture}
        }
        \caption*{(b) CHRR-Half}
    \end{minipage}
    \vspace{-3mm}
    \caption{Comparison of CHRR and CHRR-Half architectures.}
    \label{fig:architecture}
\end{figure}


\section{Experiment on XMC Datasets}
\label{sec:Realdata_experiment}
To examine the advantages of circular vectors, we conducted experiments on several XMC datasets.
Note that achieving the state-of-the-art performance on XMC datasets was not the goal of this study, which focuses on the efficiency of the learning method with circular vectors.
However, to validate the effectiveness of CHRR in the XMC task, we compared our method with several strong baselines. 
These include tree-based FastXML~\cite{Prabhu14}, PfastreXML~\cite{Prabhu18}, and deep learning based XML-CNN~\cite{zhang2018deep}, in addition to FC and HRR.

\subsection{Datasets}
\begin{table}[t]
    \centering
    \scalebox{0.8}{
        \begin{tabular}{lrrrr}
          \toprule
          Dataset   & $N_{train}$ & $N_{test}$ & $L$ & $\bar{L}$ \\ \hline
          Delicious  & 12,920 & 3,185 & 983 & 311.61 \\
          EURLex-4K & 15,539 & 3,809 & 3,993 & 25.73 \\
          Wiki10-31K & 14,146 & 6,616 & 101,938 & 8.52 \\
          Delicious-200K & 196,606 & 100,095 & 205,443 & 2.29 \\
          \bottomrule
        \end{tabular}
}
    \caption{Details of the datasets from \citet{Bhatia16}.
    Here, $N_{train}$ is the number of training samples,
    $N_{test}$ is the number of test samples,
    $L$ is the number of labels,
    $\bar{L}$ is the average number of samples per label.}
    \label{tab:datasets}
\end{table}
We evaluated our method on the four datasets for text XMC tasks from \citet{Bhatia16}.
Table~\ref{tab:datasets} shows the details of the datasets.
The features for each sample is a bag-of-words of $F$ words.

\subsection{Evaluation Metrics}
We evaluated each method by using precision at $k$ (P@$k= \frac{1}{k} \sum_{l \in \text{rank}_k(\hat{\mathbf{y}})} \mathbf{y}_l$) and 
the propensity score at $k$ (PSP@$k$),
which are commonly used metrics in the XMC task.
P@$k$ is the proportion of true labels in the top-$k$ predictions.
PSP@$k= \frac{1}{k} \sum_{l \in \text{rank}_k(\hat{\mathbf{y}})} \frac{\mathbf{y}_l}{p_l}$ is a variation of precision that takes into account the relative frequency of each label.
Here, $\text{rank}_k(\hat{\mathbf{y}})$ is the ranking of all labels in the predicted $\hat{\mathbf{y}}$ and $p_l$ is the relative frequency of the $l$-th label.
We used $k = 1, 5, 10, 20$ for P@$k$, and $k = 1, 5, 10, 20$ for PSP@$k$ in the experiments described below.

\subsection{Experimental Settings}
\label{subsec:exp_settings}
\begin{table*}[t]
    \centering
    \begin{tabular}{lrrrrrrrr}
    \toprule
    & \multicolumn{4}{c}{Delicious~($\textbf{59\%},\textbf{19\%}$)}& \multicolumn{4}{c}{Delicious-200K~($\textbf{61\%},\textbf{80\%}$)}
        \\\cmidrule(lr){2-5} \cmidrule(lr){6-9} 
        &P@1&P@5&PSP@1&PSP@5&P@1&P@5&PSP@1&PSP@5
        \\
        {FastXML} & 69.6 & \textbf{59.3} & 32.3 & 35.4 & 43.1 & 36.2 & 6.5 & 8.3
        \\
        {PfastreXML} & 67.1 & 58.6 & \textbf{34.6} & 35.9  & 41.7 & 35.6 & 3.2 & 4.4 
        \\
        {{FC}} & 70.8 & 59.2 & 34.1 & \textbf{36.1} & 35.1 & 32.1 & 5.3 & 7.4
        \\
        {{CHRR}} & \textbf{71.2} & \textbf{59.3} & 34.3 & 35.9 & \textbf{43.2} & \textbf{37.1} & \textbf{6.6} & \textbf{8.5}
        \\\cmidrule(lr){1-9}

    & \multicolumn{4}{c}{EURLex-4K~($\textbf{61\%},\textbf{99\%}$)}& \multicolumn{4}{c}{Wiki10-31K~($\textbf{62\%},\textbf{99\%}$)}
        \\\cmidrule(lr){2-5} \cmidrule(lr){6-9}
        &P@1&P@5&PSP@1&PSP@5&P@1&P@5&PSP@1&PSP@5
        \\
        {FastXML} & 76.4 & 52.0 & 33.2 & \textbf{42.0} & 83.0 & 57.8 & 9.8 & 10.5
        \\
        {PfastreXML} & 71.4 & \textbf{50.4} & 26.6 & 39.0 & 83.6 & 59.1 & \textbf{19.0}  & \textbf{18.4}
        \\
        {XML-CNN} & 75.3 & 49.2 & 32.4 & 39.5 & 81.4 & 56.1 & 9.4 & 10.2 
        \\
        {{FC}} & \textbf{77.4} & 47.9 & \textbf{33.6} & 37.3 & 80.5 & 46.4 & 10.5 & 8.9
        \\
        {{FC}$+\phi_\text{XLNet}$} & 73.3 & 48.8 & 33.0 & 40.0 & 84.0 & 58.9 & 10.9 & 11.5 
        \\
        {{CHRR}} & 75.2 & 47.8 & 28.7 & 34.9 & 82.2 & 58.8 & 10.2 & 10.9
        \\
        {{CHRR}$+\phi_\text{XLNet}$} & 77.0 & 50.0 & 29.8 & 37.6 & \textbf{86.8} & \textbf{65.1} & 11.9 & 13.0 
        \\
        \bottomrule
    \end{tabular}
    \caption{Performance comparisons of CHRR and other competing methods over four benchmark datasets, and the left number in \textbf{bold} represents the compression ratio $\left(1-\frac{(F \times h_{C} + {h_C}\times{h_C})+(h_C \times 2d + d \times L)}{(F\times h_{F} + {h_F}\times{h_F})+(h_{F} \times L )}\right)$ of the CHRR's model size for FC's model size.
    CHRR is set with $d=800$ and $h_C=768$.
    And the right number in \textbf{bold} represents the compression ratio ($1-\frac{d}{L}$) of the CHRR's output dimensions for FC's output dimensions.
    For FC, $d$ is set at the number of labels in each dataset ($L$) and $h$ is set at $2048$.
    {FC}$+\phi_\text{XLNet}$ and {CHRR}$+\phi_\text{XLNet}$  refers to the results obtained using XLNet as the feature representation.
    We obtained the results for FastXML, PfastreXML, and XML-CNN from \cite{you2019attentionxml} and \cite{yu2022pecos}.
    }
    \label{tab:results}
\end{table*}
We compared CHRR to five competitive methods~(FC, HRR, FastXML, PfastreXML, XML-CNN) over four datasets.
For the implementation of FC and HRR, we used the scripts provided by~\citet{ganesan2021learning} available at the GitHub URL.\footnote{\url{https://github.com/NeuromorphicComputationResearchProgram/Learning-with-Holographic-Reduced-Representations}}
We implemented CHRR by using PyTorch~\citep{paszke2019pytorch}.
The training methods and the model architectures basically followed the scripts provided by~\citet{ganesan2021learning}.
The learning rate was set to 1, the batch size was 64, and the number of training epochs was 100. These hyperparameters were chosen based on preliminary experiments to balance training time and model performance.
For the EURLex-4K and Wiki10-31K datasets, we also conducted experiments using both BoW~(Bag of Words) and pretrained XLNet embeddings~\cite{chang2020taming} as features. 
The dimensionality of BoW is the same as the dimensionality $F$ of the features shown in Table~\ref{tab:datasets}, and the dimensionality of XLNet as a feature is $1,024$ dimensions.
In CHRR, we varied the dimension of the symbol vectors ($d$) $\{100, 400, 800, 1000\}$.
To investigate the possibility that a larger hidden layer size $h$ improves the learning effect in FCs with large output dimensionality, we conducted experiments with three settings of hidden layer size ($h$)~$\{768, 1024, 2048\}$.
For main results, we chose $d = 800$ and $h=768$ for CHRR and $h=2048$ for FC.
All experiments are conducted with two hidden layers.

\subsection{Results and Discussion}
\label{subsec:results_and_discussion}
\begin{figure*}[ht!]
    \centering
        \begin{minipage}[b]{0.27\linewidth}         
        \centering 
        \includegraphics[width=\linewidth]
        {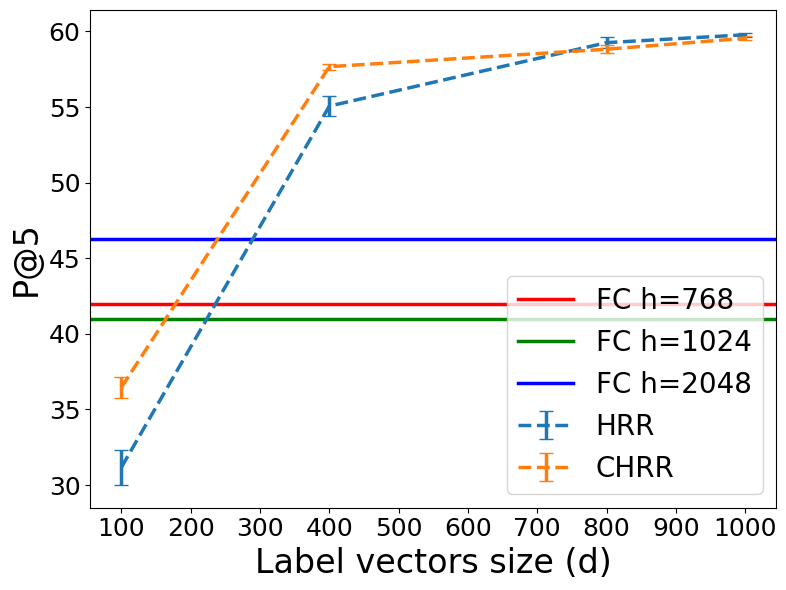}
        \vspace{-7.5mm}
        \caption*{(a)~Wiki10-31K~P@5}
        \end{minipage}
    \hfill
    \begin{minipage}[b]{0.27\linewidth}                    
        \centering
        \includegraphics[width=\linewidth]
        {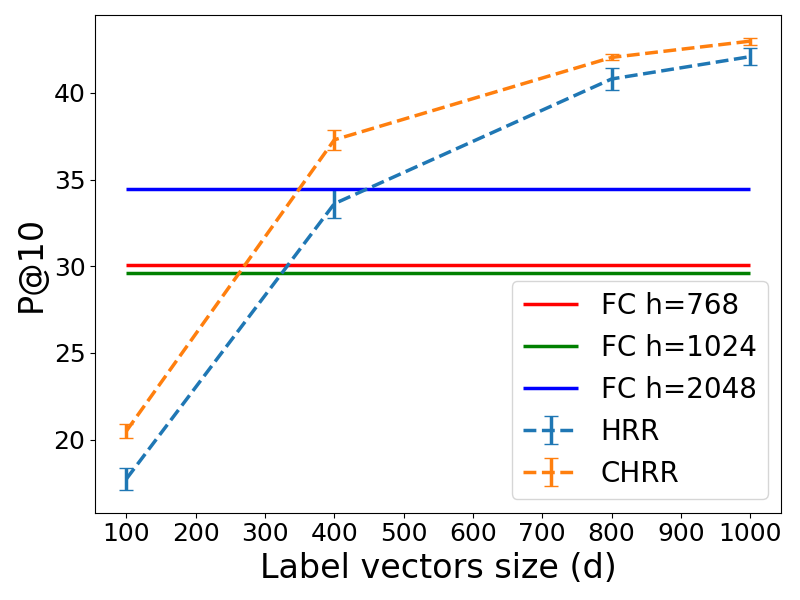}
        \vspace{-7.5mm}
        \caption*{(b)~Wiki10-31K~P@10}
    \end{minipage}
    \hfill
    \begin{minipage}[b]{0.27\linewidth}        
        \centering
        \includegraphics[width=\linewidth]
        {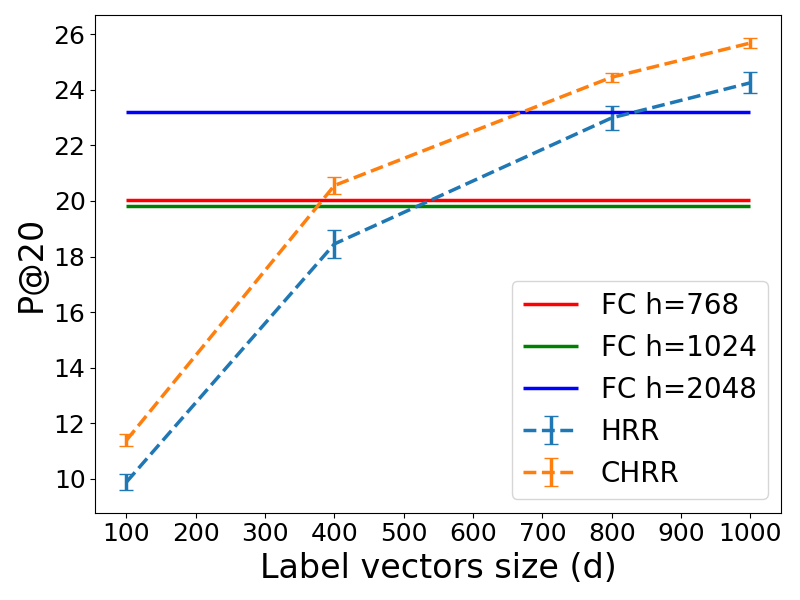}
        \vspace{-7.5mm}   
        \caption*{(c)~Wiki10-31K~P@20}
    \end{minipage}
        \begin{minipage}[b]{0.27\linewidth}         
        \centering 
        \includegraphics[width=\linewidth]   {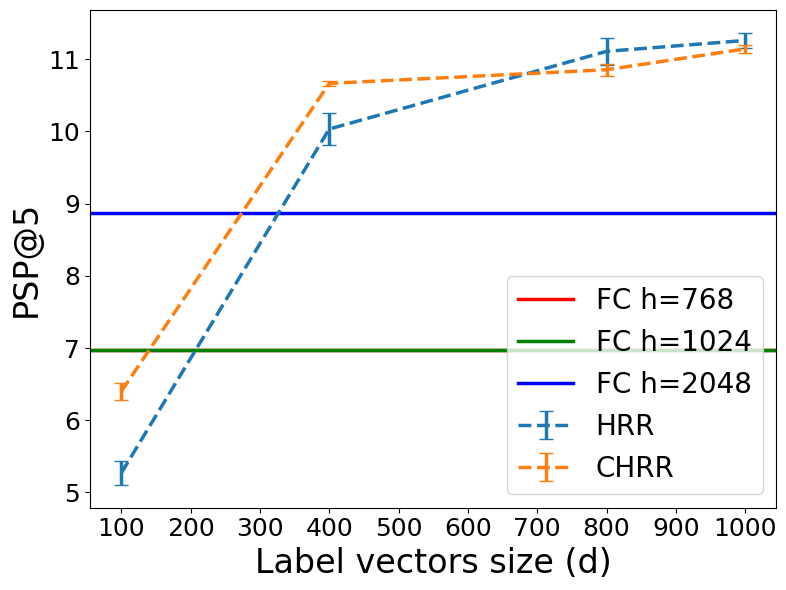}
        \vspace{-7.5mm}
        \caption*{(d)~Wiki10-31K~PSP@5}
    \end{minipage}
    \hfill
    \begin{minipage}[b]{0.27\linewidth}                    
        \centering
        \includegraphics[width=\linewidth]
        {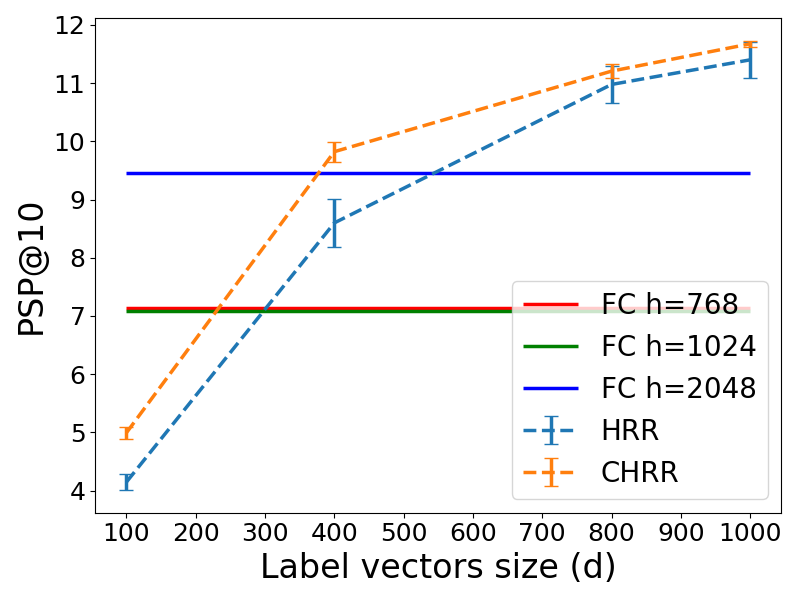}
        \vspace{-7.5mm}
        \caption*{(e)~Wiki10-31K~PSP@10}
    \end{minipage}
    \hfill
    \begin{minipage}[b]{0.27\linewidth}        
        \centering
        \includegraphics[width=\linewidth]        {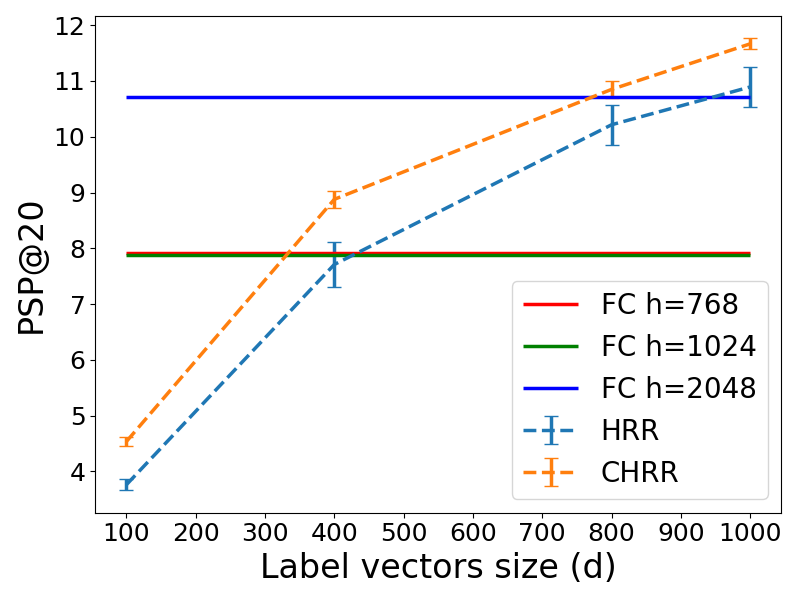}
        \vspace{-7.5mm}   
        \caption*{(f)~Wiki10-31K~PSP@20}
    \end{minipage}
        \begin{minipage}[b]{0.27\linewidth}         
        \centering 
        \includegraphics[width=\linewidth]{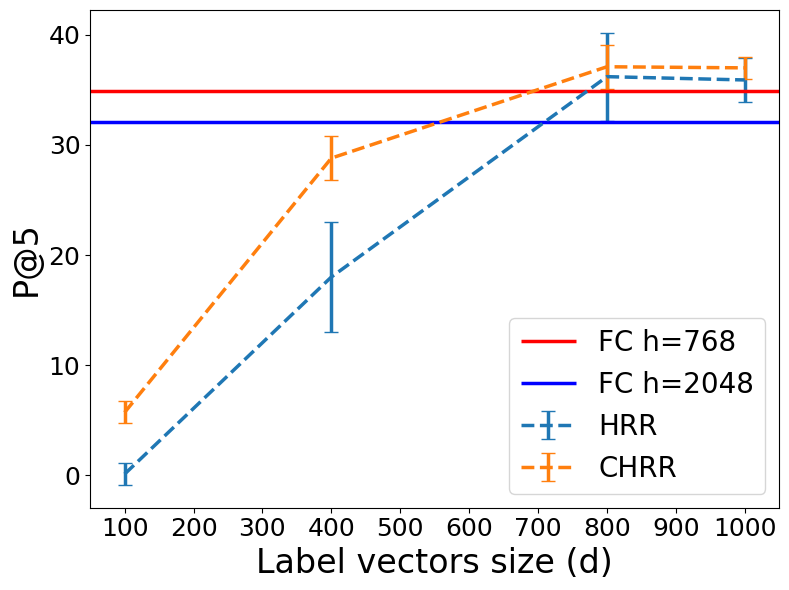}
        \vspace{-7.5mm}
        \caption*{(g)~Delicious-200K~P@5}
    \end{minipage}
    \hfill
    \begin{minipage}[b]{0.27\linewidth}                    
        \centering
        \includegraphics[width=\linewidth]{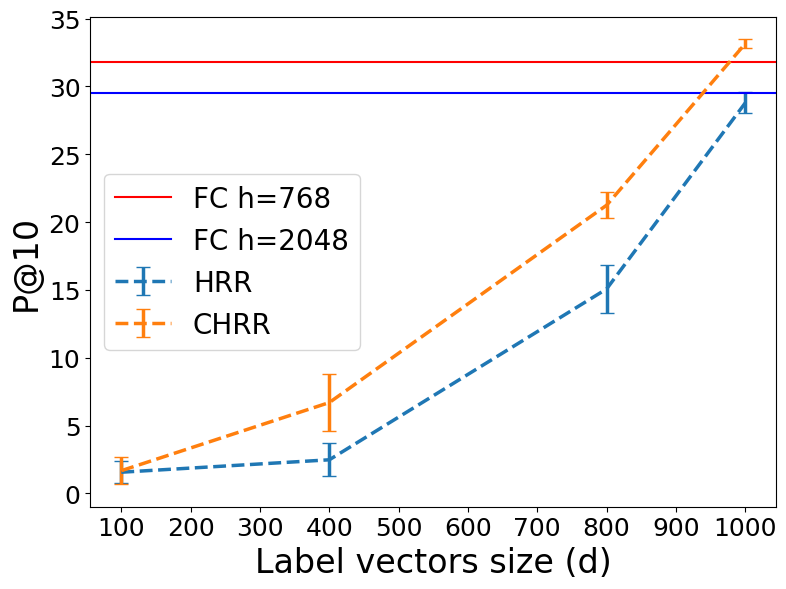}
        \vspace{-7.5mm}
        \caption*{(h)~Delicious-200K~P@10}
    \end{minipage}
    \hfill
    \begin{minipage}[b]{0.27\linewidth}        
        \centering
        \includegraphics[width=\linewidth]{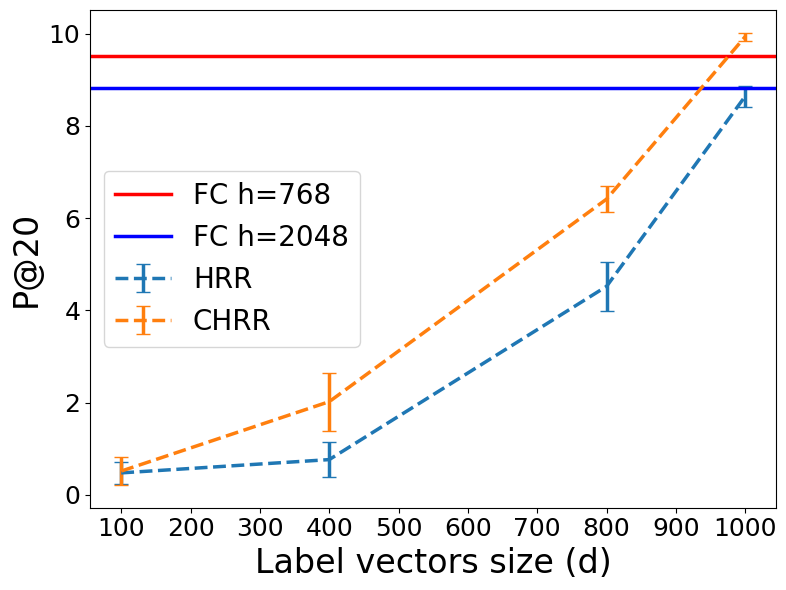}
        \vspace{-7.5mm}  
        \caption*{(i)~Delicious-200K~P@20}
    \end{minipage}
        \begin{minipage}[b]{0.27\linewidth}         
        \centering 
        \includegraphics[width=\linewidth]{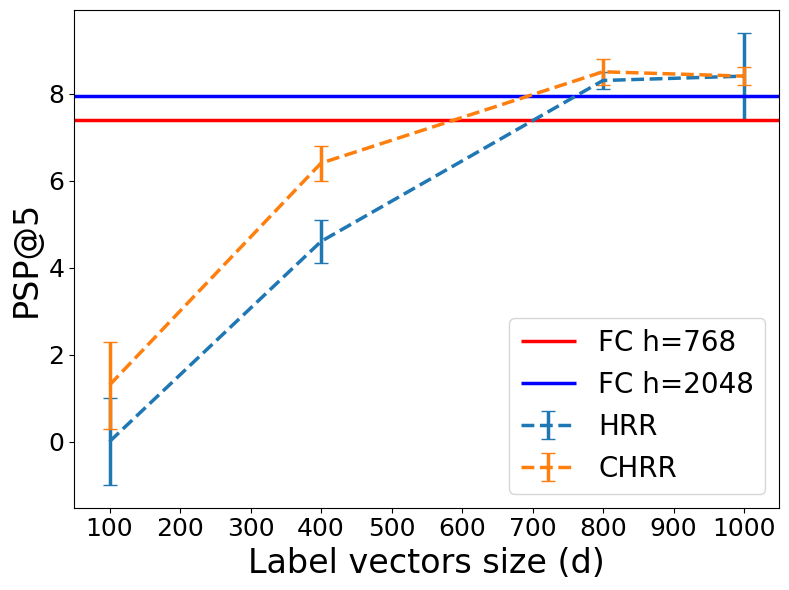}
        \vspace{-7.5mm}
        \caption*{(j)~Delicious-200K~PSP@5}
    \end{minipage}
    \hfill
    \begin{minipage}[b]{0.27\linewidth}                    
        \centering
        \includegraphics[width=\linewidth]{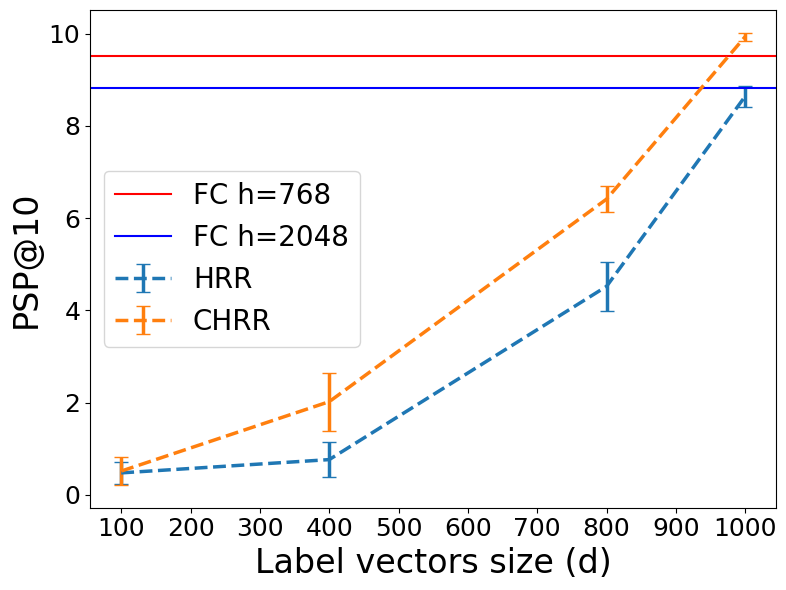}
        \vspace{-7.5mm}
        \caption*{(k)~Delicious-200K~PSP@10}
    \end{minipage}
    \hfill
    \begin{minipage}[b]{0.27\linewidth}        
        \centering
        \includegraphics[width=\linewidth]{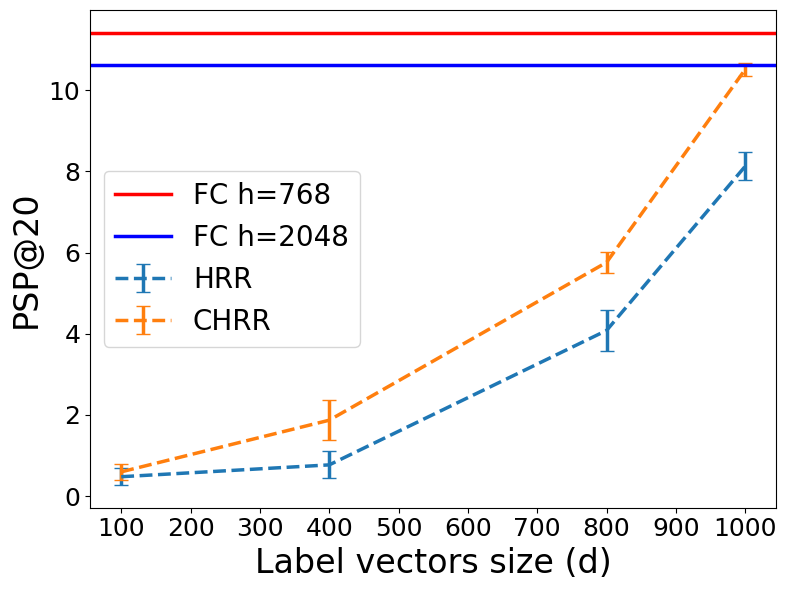}
        \vspace{-7.5mm} 
        \caption*{(l)~Delicious-200K~PSP@20}
    \end{minipage}
    \vspace{-3mm} 
    \caption{Impact of the number of dimensions ($d$) on P@5, P@10, P@20, PSP@5, PSP@10, and PSP@20 for Wiki10-31K and Delicious-200K datasets.
    We used BoW as features in all models.}
    \label{fig:results_figure}
\end{figure*}
Table~\ref{tab:results} lists P@1, P@5, PSP@1, and PSP@5 for the CHRR model, with five standard methods.
CHRR achieves up to 99\% output dimension compression and 62\% model size reduction compared to FC, which is comparable or better than other baselines.
CHRR$+\phi_\text{XLNet}$ with XLNet as a feature showed higher results than the CHRR case with BoW. 
In particular, it showed significant improvement on the Wiki10-31K dataset.
Figure~\ref{fig:results_figure} shows the impact of the dimensionality size $d$ of the HRR and CHRR on performance, in addition to the FC results.
On certain datasets, CHRR outperformed FC even when it had vectors with lower dimensions.
These results suggest that CHRR has a higher capacity for learning on datasets with a large number of labels than FC does.

We also compared CHRR with HRR.
As shown in Figure~\ref{fig:results_figure}, CHRR was better than HRR in many cases.
In particular, the results for P@20 and PSP@20, where the value of the evaluation index k is large, we confirmed that the difference in performance is significant.
As our theoretical experiment in~\S~\ref{subsec:Retrieval_experiment} showed, CHRR could represent many labels with high accuracy even for low-dimensional vectors. 
The results of the theoretical experiments in \S~\ref{subsec:Retrieval_experiment} and the experiment on real datasets in \S~\ref{sec:Realdata_experiment} suggest that the CHRR is able to represent a larger number of correct labels.

\subsection{Impact of Model Architecture}
\begin{figure*}[h!]
    \centering
    \begin{minipage}[b]{0.27\linewidth}
        \centering
        \includegraphics[width=\linewidth]{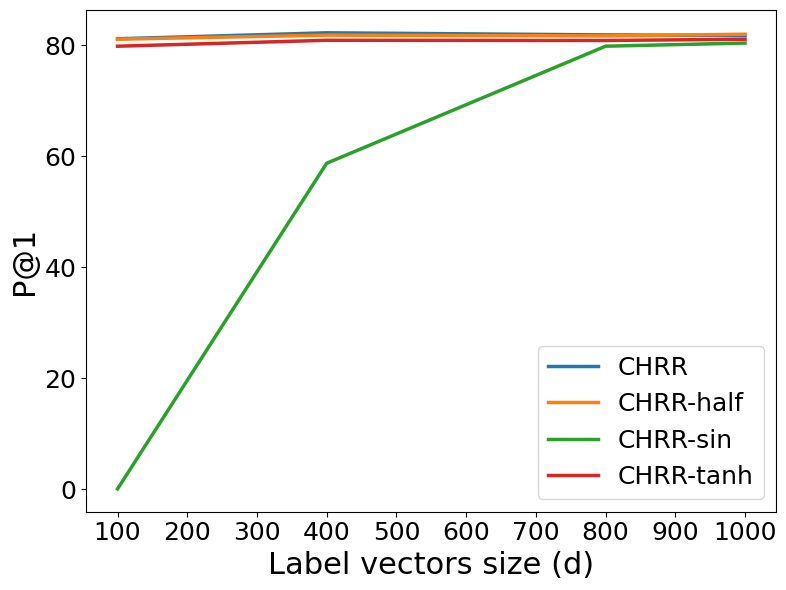}
        \vspace{-7mm}
        \caption*{(a) P@1}
    \end{minipage}
    \hfill
    \begin{minipage}[b]{0.27\linewidth}
        \centering
        \includegraphics[width=\linewidth]{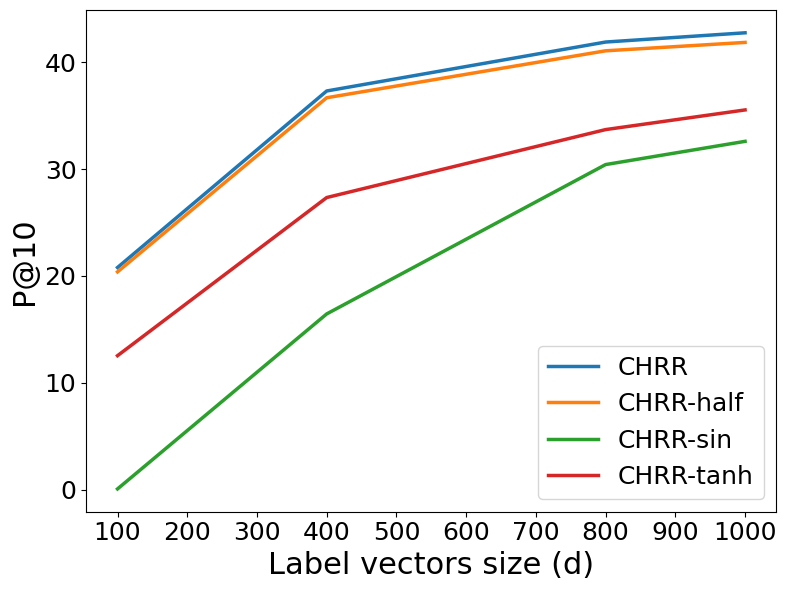}
        \vspace{-7mm}
        \caption*{(b) P@10}
    \end{minipage}
    \hfill
    \begin{minipage}[b]{0.27\linewidth}
        \centering
        \includegraphics[width=\linewidth]{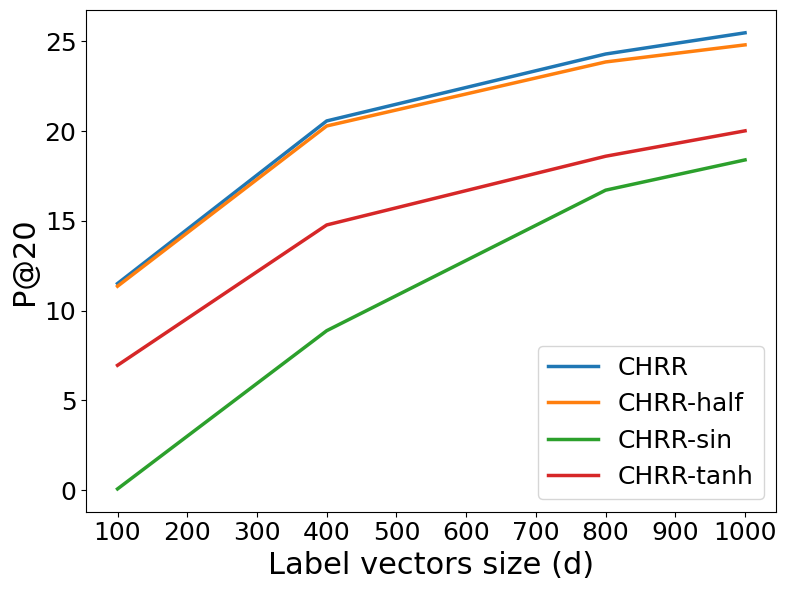}
        \vspace{-7mm}
        \caption*{(c) P@20}
    \end{minipage}
    \vspace{-3mm} 
    \caption{Comparison of CHRR variants (CHRR, -Half, -sin, and -tanh) on the Wiki10-31K dataset.}
    \label{fig:Model_diff}
\end{figure*}
This section describes the results of the experiments on the impact of the model architectures in \S~\ref{subsec:model_ablation}.
Figure~\ref{fig:Model_diff} compares the performances of the CHRR variants (CHRR, CHRR-Half, CHRR-sin, and CHRR-tanh) on the Wiki10-31K dataset.
As mentioned in \S~\ref{subsec:model_ablation}, there was no significant difference in performance between CHRR and this model.
CHRR-sin and CHRR-tanh both obtained similar results that were inferior to those of CHRR and CHRR-Half. 
While the $\sin$ function in CHRR-sin seems to consider the cyclic feature, the results show that it is imperfect at predicting the of the circular-label vector.
In short, our developed network architecture is important for the XMC learning with circular vectors, while the increase in the model size is not a big issue.

\section{Conclusion}
\label{sec:conclusion}
The XMC task still faces challenge of dealing with a large number of output labels.
In this paper, we attempted to address this issue by using a low dimensional circular vector to output directly.
In theoretical experiments in \S~\ref{subsec:Retrieval_experiment} and \S~\ref{subsec:Variance_experiment}, we showed that many labels can be accurately encoded by using circular vectors (CHRR) rather than normal real-valued vectors (HRR).
Moreover, using actual XMC datasets,
we compared CHRR with baseline methods in \S~\ref{sec:Realdata_experiment}. 
CHRR reduced the output layer size by up to 99\% compared to FC, while it outperformed other baselines in most results.
Comparing HRR and CHRR, CHRR outperformed on most results.
In the future, we will incorporate circular vector systems into other DNN models such as LSTM~\citep{hochreiter1997long} and Transformer~\citep{vaswani2017attention},
as well as Associative LSTM~\citep{DBLP:journals/corr/DanihelkaWUKG16} and Hrrformer~\citep{alam2023recasting}.

\clearpage
\section*{Limitations}

Our study has several limitations that should be considered in interpreting the results:

\begin{enumerate}
    \item \textbf{Model Age and Adaptability:} The HRR and CHRR models utilized in our experiments are based on established frameworks that may not incorporate the latest advancements in neural network architectures \cite{ganesan2021learning}. Newer models or hybrid approaches might offer improved performance.
    
    \item \textbf{Comparison with State-of-the-Art XMC Models:} Our study did not include a comparison with the latest models in the Extreme Multi-label Classification (XMC) domain, such as APLC-XLNet \cite{ye2020pretrainedgeneralizedautoregressivemodel}, LightXML \cite{jiang2021lightxml}, AttentionXML \cite{you2019attentionxml}, and CascadeXML \cite{kharbanda2022cascadexmlrethinkingtransformersendtoend}. Future research should consider comparing the performance of HRR, HRR(w/Proj), and CHRR against these state-of-the-art models to provide a more comprehensive evaluation of their effectiveness in .

    \item \textbf{Comparison with LLM:} Our study did not include a comparison with the Large Language Model (LLM) approach, which is currently the state-of-the-art in various NLP tasks. Future research should consider comparing the performance of HRR, HRR(w/Proj), and CHRR against LLMs to further evaluate the effectiveness of these models.
\end{enumerate}

These limitations underscore the need for further research to refine and extend the applicability of the models proposed in this study.

\section*{Ethics Statement}

We used the publicly available XMC datasets, Delicious, EURLex-4K, Wiki10-31K and Delicious-200K, to train and evaluate DNN models, and there is no ethical consideration.

\section*{Reproducibility Statement}
As mentioned in \S~\ref{subsec:exp_settings}, we used the publicly available code to implement FC, HRR and CHRR.
Our code will be available at \url{https://github.com/Nishiken1/Circular-HRR}.


\bibliography{anthology,custom}
\bibliographystyle{acl_natbib}

\appendix

\end{document}